\documentclass[conference, a4paper]{IEEEtran}

\usepackage{times}
\usepackage{graphicx}
\usepackage{amsmath}   
\usepackage{amssymb}
\usepackage{epstopdf}
\usepackage{url}
\usepackage{subfigure}
\usepackage{enumitem}
\usepackage{algorithm} 
\usepackage{algpseudocode}
\DeclareMathOperator*{\argmin}{arg\,min}
\usepackage{subfigure}

\renewcommand\IEEEkeywordsname{Keywords}

\begin{document}

\title{PATH: Person Authentication using Trace Histories}

\author{Upal Mahbub \quad Rama Chellappa\\
Department of Electrical and Computer Engineering and the Center for Automation Research, \\UMIACS, University of Maryland, College Park, MD 20742\\
{\tt\small \{umahbub,rama\}@umiacs.umd.edu}}

\maketitle

\begin{abstract}
In this paper, a solution to the problem of Active Authentication using trace histories is addressed. Specifically, the task is to perform user verification on mobile devices using historical location traces of the user as a function of time. Considering the movement of a human as a Markovian motion, a modified Hidden Markov Model (HMM)-based solution is proposed. The proposed method, namely the Marginally Smoothed HMM (MSHMM), utilizes the marginal probabilities of location and timing information of the observations to smooth-out the emission probabilities while training. Hence, it can efficiently handle unforeseen observations during the test phase. The verification performance of this method is compared to a sequence matching (SM) method , a Markov Chain-based method (MC) and an HMM with basic Laplace Smoothing (HMM-lap). Experimental results using the location information of the UMD Active Authentication Dataset-02 (UMDAA02) and the GeoLife dataset are presented. The proposed MSHMM method outperforms the compared methods in terms of equal error rate (EER). Additionally, the effects of different parameters on the proposed method are discussed.

\end{abstract}

\begin{IEEEkeywords}
Active authentication; geo-location-based verification; hidden markov models
\end{IEEEkeywords}

\section{Introduction}
Recognizing human behavior and understanding user mobility from sensor data is an interesting and challenging problem in ubiquitous computing \cite{understanding-mobility-based-on-gps-data}, \cite{Understanding_Transportation_Modes_Based_on_GPS_Data_for_Web_Applications}. From mobility information of a user, it is possible to recognize social patterns in daily activities, identify socially significant locations, and model organizational rhythms \cite{MITRealityDataset}. Recent proliferation of mobile devices like smartphones and tablets has made it possible to collect mobility data precisely and easily for an extended period of time and therefore research activity in mining the mobility data to infer the pattern of life is increasing \cite{mining-individual-life-pattern-based-on-location-history}. On the other hand, security concerns about personal information stored in mobile devices is on the rise too because of their sensitive nature \cite{PhoneSecurity_Chin}. Industry surveys estimate that  $34\%$ of smartphone users in the U.S. do not lock their phones with passwords \cite{ConsumerReport:PhoneSecurity}, \cite{SmartPhoneNotSmartEnough:Fischer}, mostly, due to the time-consuming, cumbersome and error-prone hassles of entering passwords on virtual keyboards or due to users' beliefs that extra passwords are not needed \cite{SmartPhoneNotSmartEnough:Fischer}. Since it is difficult to remember and type stronger passwords $150$ times per day \cite{MeekerReport2013}, which is the average number of cellphone access per user, weaker passwords are still preferred by the users and they become victims of attacks on smart phones $76\%$ of times \cite{VerizonReportDBIR}.

In this regard, the concept of Active Authentication (AA) has emerged recently, in which the enrolled user is authenticated continuously in the background based on the user's biometrics such as faces captured by the front-camera \cite{AA_Samangouei}, \cite{AA_Fathy}, \cite{FSFD_Mahbub}, touch screen gesture \cite{TouchAA_Feng}, \cite{umd_Dataset}, typing pattern \cite{TypingAA_Arujo} etc. When a person uses the phone, the AA system compares the usage pattern with the enrolled user's pattern of use and either deems that the usage patterns are sufficiently similar and make the full functionality of the phone (including sensitive applications and data) available, or it blocks the current user from accessing anything \cite{AA02_MahbubChellappa_BTAS2016}. 

Combining the usability of location traces to model a user's pattern of life with the concept of active authentication, the Person Authentication using Trace Histories (PATH) problem is addressed. In PATH, the goal is to perform user verification from historical location data of a user in a continuous manner so that a verification score based on the location information is obtained continuously. This score can be fused with scores returned by other modalities such as touch or face to improve the performance of the overall authentication system. 

The key motivation of this paper is the wide availability of individual GPS data through smartphones, wearable devices etc. and the discriminating life patterns of different individuals. The goal of this research is to generate a confidence score for authenticating the current user of a smartphone using recent location traces based on historical trace data of the original user. The contributions of this paper are:
\begin{itemize}
\item A novel formulation of the user verification problem using historical location traces is proposed. Confidence scores from location traces for the purpose of authentication, taking into consideration the sparseness of the geo-location data are produced.
\item A unique method for user data clustering for trace state generation which is capable of handling a large number of unknown locations is introduced.
\item In order to account for unforeseen observations during testing phase, a modified Hidden Markov Model(HMM)-based user verification method, namely, Marginally Smoothed HMM (MSHMM) is proposed.
\end{itemize}

The rest of the paper is organized as follows. In Section II, background and related works on this topic are discussed. In Section III, the PATH problem is explained in detail along with the difficulties and possible solutions. Different solutions to the PATH problem are described in Section IV and experimental results and discussions are presented in Section V. Finally, conclusion and suggestions for future work are given in Section VI.

\section{Related Works}
\subsection{Active Authentication Techniques}
Faces \cite{AA_Fathy}, \cite{AA_Samangouei}, \cite{FSFD_Mahbub}, touch/swipe signatures \cite{Touch_SerwaddaPW13},\cite{Heng_WACV2015}, \cite{umd_Dataset}, gait \cite{AA_Gait} and device movement-patterns/accelerometer  \cite{AccDataAuth_Primo2014}, \cite{ContAuth_AJain} are the most explored modalities for active authentication. Face-based authentication, though most accurate, requires more computational power and drains battery faster, whereas, swipe and accelerometer are less computationally expensive but are not discriminative enough. Among the other AA approaches, in \cite{AA_StylometryAppWeb_Friedman}, the authors fused stylometry with application usage, web browsing data and location information, and, in \cite{Heng_FG2015_Fusion} the authors fused face and touch gestures for AA. Most of the methods applied for AA perform feature extraction from training samples and generate matching scores based on cosine or Euclidean distances in the verification setup \cite{VMP_SPM_AA_2016}.

\subsection{Location History Mining}
Purely location-based verification approaches for AA are yet to be found in the literature. Instead, most location-based research reports are focused on data mining to obtain information about an individual's pattern of life. For example, in \cite{mining-individual-life-pattern-based-on-location-history} the authors predict the user’s movement among the location points and infer user-specific activity at each location. In \cite{MMC_NextPlace}, the authors focus on detecting significant locations of a user and predicting the user's next location or infering the daily movements. In \cite{Understanding_Transportation_Modes_Based_on_GPS_Data_for_Web_Applications}, \cite{HighLevelBehavior_Patterson2003}  and \cite{understanding-mobility-based-on-gps-data}, the authors infer the high-level behavior of the user, such as, the transportation modes on the way to the point locations. Other research efforts on GPS location data include driving behavior mining, finding mode of transportation and the most likely route, learing a Bayesian model of travel through an urban environment etc. \cite{mining-individual-life-pattern-based-on-location-history}.

\subsection{Geo-location Data Processing}
When mining individual life patterns from geo-location data, the problem can be considered as a sequential pattern mining problem, widely used in health-care data processing, web usage analysis, text mining for natural language processing, speech processing, sequential image processing, bioinformatics and in many other domains \cite{applications-for-pattern-discovery-using-sequential-data-mining}. In \cite{mining-individual-life-pattern-based-on-location-history}, sequential pattern mining has been employed for individual life pattern modeling. Since, the geo-location trajectory data are spatio-temporal in nature, the fuzziness of space (usually no two point in the trajectory data are exactly the same) prevents the direct use of traditional frequent pattern mining algorithms. The usual practice is to cluster the geo-location points onto finite number of observation states and then perform sequential pattern mining on the state transition trajectories. In natural language processing, template matching approaches have been employed for matching features from a text sequence with pre-calculated feature vectors using edit distance or some other distance metric. String matching algorithms are also found to be effective in this regard. A different type of approach is based on building state-space models like a Markov Chain or a hidden markov model, from temporal data. Several research works on next place prediction from location history, such as in \cite{MMC_NextPlace}, \cite{MMC_Showme_Gambs}, \cite{HMMLocationPredict_Mathew}, are based on state-space models like Mobility Markov Chains (MMC), Mixed Markov Chain (MMM) and Hidden Markov Models (HMM).

\subsection{Dataset of Smarphone Location Service \\Records}
The Geolife GPS trajectory dataset was collected in Microsoft Research Asia by 182 users over four years (from April 2007 to October 2011). The GPS trajectories of this dataset are represented by sequences of time-stamped points, each containing latitude, longitude and altitude information. The dataset contains $17,621$ trajectories with a total distance of $1,251,654$ kilometers and a total duration of $48,203$ hours. These trajectories were recorded by different GPS loggers and GPS-phones, and have a variety of sampling rates. $91$ percent of the trajectories are logged in a dense representation, e.g. every $1 \sim 5$ seconds or every $5 \sim 10$ meters per point \cite{Geolife_CooperativeSocialNetworkingService}, \cite{geolife-gps-trajectory-dataset-user-guide}. Apart from the GPS trajectories, the dataset contains information about a broad range of users' outdoor movements, including not only life routines like going to work or home, shopping, hiking etc. but also some entertainments and sports activities, such as shopping, sightseeing, dining, hiking, and cycling. This trajectory dataset has been used in many research fields, such as mobility pattern mining, user activity recognition, location-based social networks, location privacy, and location recommendation \cite{Understanding_Transportation_Modes_Based_on_GPS_Data_for_Web_Applications}, \cite{understanding-mobility-based-on-gps-data}, \cite{mining-individual-life-pattern-based-on-location-history}.

The largest known dataset on smartphone usage is the Google's Project Abacus data set consisting of $27.62$ TB of smartphone sensor signals collected passively from approximately $1500$ users for six months on Nexus $5$ phones \cite{NataliaNeverova_Abacus}. Apart from location service data, this dataset also contains images from front-facing camera and data from touchscreen and keyboard, gyroscope, accelerometer, magnetometer, ambient light sensor, Bluetooth, WiFi, cell antennae, app usage etc. with time statistics. However, this dataset is not available for the research community due to privacy issues.

\begin{table}
\small
\centering
\caption{General Information on Geo-location Data}
\begin{tabular}{p{6.0cm} c}
\hline
No. of Subjects	                					&$45$\\
\hline
Avg. No. of Sessions/User with Location Data    	&$\backsim$ $186$\\
\hline
Total Number of Location Traces & $8303813$ \\
\hline
Number of Location Traces Per User			&$\backsim$  $184529$ \\
\hline
Number of Location Traces Per Session			& $\backsim$ $993$ \\\hline
\end{tabular}
\label{GeoData}
\vskip -5pt
\end{table}

Recently, the UMDAA02 data set has been published which contains $141.14$ GB of smartphone sensor signals collected passively from $48$ volunteers on Nexus 5 phones over a period of 2 months \cite{AA02_MahbubChellappa_BTAS2016}. The sensors from which data was collected include the front-facing camera, touchscreen, gyroscope, accelerometer, magnetometer, ambient light sensor, location service, Bluetooth, WiFi, cell antenna, proximity sensor, temperature sensor and pressure sensor. The data collection application also stored the timing of screen lock and unlock events, start and end time stamps of calls, currently running foreground application etc. The volunteers, who used the data collection phone as their primary device for a week, were given the option to stop data collection at will and review the stored data prior to sharing it for research purposes. The dataset contains geo-location data obtained from $45$ users (summarized in Table \ref{GeoData}).

\section{Problem Formulation}
\begin{figure}[t]
\centering
\includegraphics[width=0.35\textwidth]{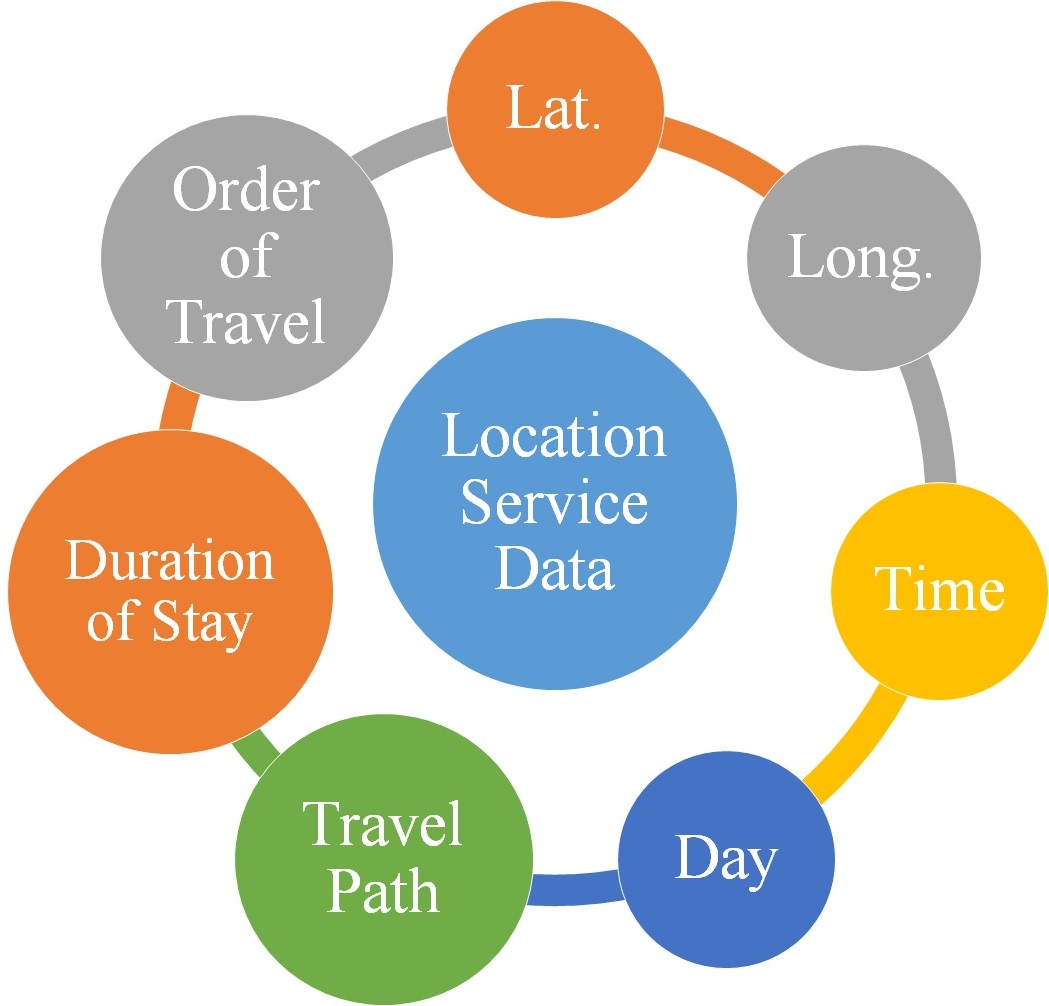}
\vskip 0pt
\caption{Useful information obtained from the location service of the smartphone.}
\label{LocationServiceInfo}
\vskip -10pt
\end{figure}

\begin{figure*}[t]
\centering
\includegraphics[width=0.8\textwidth]{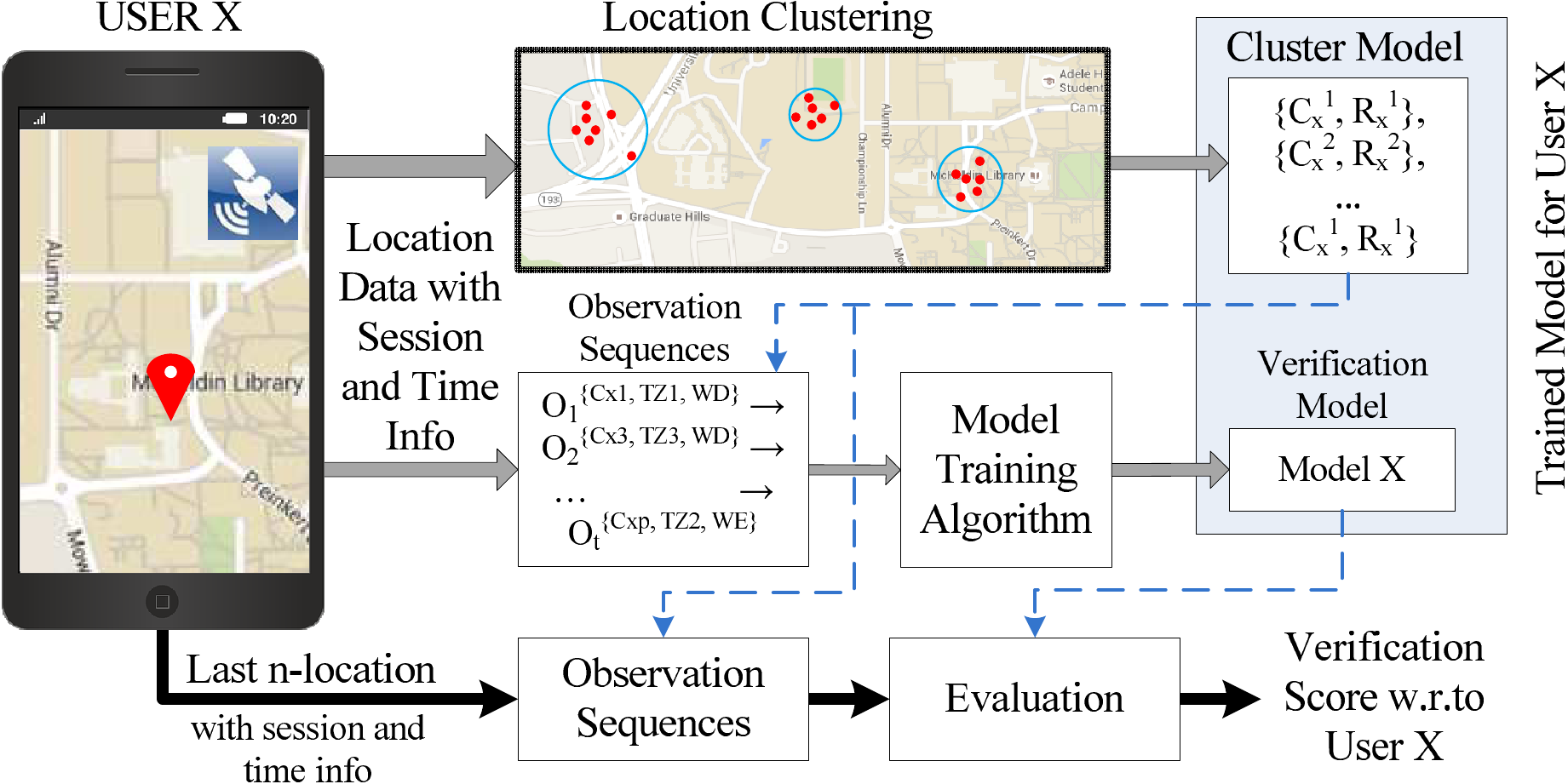}
\vskip 0pt
\caption{System overview for handling the PATH problem.}
\label{VeriSteps}
\vskip -0pt
\end{figure*}

The location service of smartphones returns geographical location of the user based on GPS and WiFi network. In addition to the latitude and longitude information, the exact day, time and duration of being in a proximity can be extracted from the location service data (Fig. \ref{LocationServiceInfo}) which are very useful for modeling the pattern of a user's location trace.

In general, the PATH problem has three challenges:
\begin{enumerate}
\item Clustering of Geo-location points to form observation states taking into account the temporal information.
\item  Handling unforeseen observation states and learning a model for each user using the sequential patterns inferred from the observation states .
\item Generating verification score from a test sequence using the trained model.
\end{enumerate}

In Fig. \ref{VeriSteps}, a schematic of the proposed verification system is shown. Basically, there are three steps. First, from the training geo-location data of user $x$, location clusters are formed and the cluster centers and radius are extracted. Then, from the sequence of geo-location data with time-stamps and session information, the sequence of training observations is obtained which is used to train the verification model for user $x$. Finally, from every $n$ observations in the test data, scores are generated to verify user $x$ using the previously trained model.

\subsection{Geo-location Points to Observation States}
\begin{figure*}[t]
\centering
\includegraphics[width=0.8\textwidth]{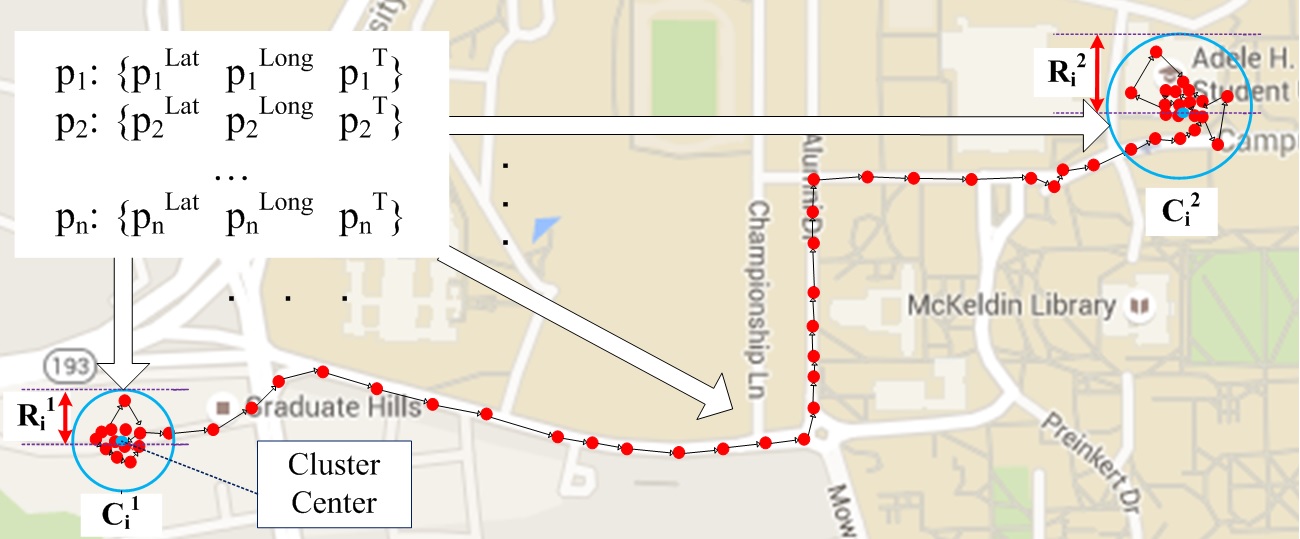}
\vskip 0pt
\caption{Geo-location Points and Clusters.}
\label{GPSClustering}
\vskip -15pt
\end{figure*}

Since, this is a verification problem as opposed to recognition, only the information about the legitimate user is available during training. For a user, the data collected by the location service is a sequence of geo-location points $P=\{p_1, p_2, \hdots, p_n\}$, where, each point $p_i \in P$ contains the longitude ($p_i^{Long}$), latitude ($p_i^{Lat}$) and time stamp ($p_i^{T}$). These points are sampled at variable rates based on the speed of movement and therefore the points might not be equally spaced in time. A location trace can be formed by connecting the Geo-location points according to their time series as shown with the red points connected with arrows in Fig. \ref{GPSClustering}.

In order to take into account the duration of stay in a certain locality, the geo-location traces are sampled once every three minutes. Thus, if a user is in the same location for a while, the location logger will log the same location at three minute interval. The historical Geo-location points of the user obtained this way are clustered into $\eta=1, \hdots, N$ clusters, namely $C^1 \hdots C^{N}$, using the DBSCAN algorithm \cite{DBScanClustering} based on geographical distances (GeoDist) between data points. The maximum distance between a point from the center of the cluster in which that point belongs is set to be below a certain value $R_{max}$ meters. A cluster $C^j$ is completely defined by $\{C^j, R^j\}$, where $C^j: \{c^j_{Lat}, c^j_{Long}\}$, $j \in \eta$, consists of the latitude and longitude of the cluster center, respectively, and $R^j$ is the radius of the cluster where $R^j \leq R_{max}$. An example of such clustering is shown in Fig. \ref{GPSClustering}), where the cluster $C^1_i$ represents presence in a residential area and $C^2_i$ in an office building in an university imply plausible regions like home, university etc that the user $i$ would visit. 

Two types of additional clusters, Transit ($Tr$) and Unknown ($Unk$), are assigned for each user. If the user is traveling, causing location information to change rapidly ($\geq 2ms^{-1}$), then those geo-locations points are assigned to $Tr$. For each of the other data points $p_m$ that are not inside any location cluster and also not in the Transit cluster, the nearest known location cluster to each of those points within a radius of $M = 10000$ meters is determined by calculating $GeoDist(j,m)$, the geological distance between $p_m$ and all $C^j$s. Then those data points are assigned to the cluster $Unk_{\ell}$, where, 
\begin{equation}
\ell=\begin{cases}
	\argmin_{j} GeoDist(j, m) , & \text{if $GeoDist(j,m)\leq M$}\\
	\infty, & \text{otherwise}
	\end{cases}
\end{equation} 
for $m=1, \hdots, n$, $p_m \notin C^j$, $j \in \eta$, $p_m \notin Tr$. 

So, to summarize, there would be $N$ location clusters $C^{j}$ and $N$ corresponding nearby unknown clusters $Unk_{j}$ for $j\in \eta$, one more cluster denoted as $Unk_{\inf}$ and the transit cluster $Tr$ totaling the number of cluster to be $2N+2$. Data points at each cluster are assigned to six different observations based on day and time information. Weekdays and weekend data points are flagged with $WD$ and $WE$. Also, the whole day is divided into three time zones (TZs) - TZ1 (12:01 am to 8:00 am), TZ2 (8:01 am to 4:00 pm) and TZ3 (4:01 pm to 12:00 pm). Thus, there are $(2N+2)\times 2\times 3$ possible observation states for the locations, transition and unknowns. One additional observation state, namely, $Null$, is considered in order to take the sparsity of the data into account. The $Null$ is inserted at the end of each day. It signifies the unavailability of location, time zone, number of observation samples and observation states in between consecutive sessions occurring in two different days.

Now, it is possible that many of the observation states are not present in the training data and yet, they may appear in the test
data because of the following reasons
\begin{enumerate}
\item The location service of the phone might only collect data when the phone is turned on and in use. Also, some user prefer to turn off the location service when the battery is low. 
\item Some unknown states near known locations might not occur during the training phase.
\item Data for all time zones and days might not be present for all location.
\end{enumerate}

\subsection{Handling Unforeseen Observations to Learn User Models}
In order to verify the user, it is imperative to take the unforeseen observations into account rather them assigning zero emission probability to them. Fortunately, the probability distribution of the occurrence of all the states can be smoothed out using the estimated values from the training data. Laplace smoothing is one easy choice to make sure that the probability does not go to zero at all, however, it does not take into account the prior information about available states. For example, assume that the observation $C^{1}-TZ1-WE$ is not present in the training set, i.e. the historical location log of the user does not contain any information of the user being at $C^1$ during timezone $1$ on weekends. The prior probability of its occurrence $P(C^{1}-TZ1-WE)$ can be still be approximated from the probability of the user being in $C^1$ during $TZ1$ and the probability of the user being in $C^1$ during weekends by assuming that the two events are independent. In the next section, the integration of this assumption into the proposed HMM training model is elaborated.

From the temporally sorted training observation sequence, three different approaches are presented for user verification. The approaches are:
\begin{enumerate}
\item Simple time-sequence matching
\item Markov Chain Models
\item MSHMM models - a Hidden Markov Model with the proposed marginal smoothing
\end{enumerate}
These approaches are discussed in details in the next section.

\section{User Verification Methods}
Three different user verification methods are discussed here. After the pre-processing step, the observations are available in a time sequences for each user and user-wise models are generated using this data.
 
\subsection{Sequence Matching (SM) Method}

\begin{algorithm}
\caption{Sequence Matching Algorithm}\label{SeqMatchAlg}
\begin{algorithmic}
\Procedure{SeqMatching}{$S_{tr}^{i},S_{te}^j$}\Comment{Training Sequence Vector of user $i$ ($S_{tr}^{i}$), n-last Test Sequence Vector of user $j$ ($S_{te}^j$)}
\State $S_c \gets 0$\Comment{Sequence Counter}
\State $S_t \gets 1$\Comment{Sequence Track Variable}
\For{$v_{tr} \in S_{tr}^{i}$}
	\If {$v_{tr} == S_{te}^j[S_t]$}
  		\State $S_t \gets S_c+1$\Comment{Element Matched}
  		\If{$St==|S_{te}^j|$}
 			\State $S_t \gets 0$
 			\State $S_c \gets S_c+1$
 		\EndIf
 	\EndIf
\EndFor

\State $r \gets (S_c+1.0)|S_{te}^j|+\frac{S_t}{|S_{tr}^i|+|S_{te}^j|)}$
\State \textbf{return} $r$\Comment{The match ratio is $r$}
\EndProcedure
\end{algorithmic}
\end{algorithm}

\subsection{Markov Chain (MC)-Based Verification}
For Markov Chain-based verification, the probability of moving to a state depends only on the last visited state and the transition matrix for all probable states.  The model $X_n$ is a Markov chain for observation sequences of length $n$ which is composed of a set of $k$-observation states $S={s_1, s_2, \hdots, s_k}$, prior probability $\rho_i=Prob\{X_0=i\}$ of entering state $i$,  and a set of transitions $t_{i,j}$ where 
\begin{equation}
t_{i,j}=Prob(X_n=s_j|X_{n-1}=s_i).
\end{equation}
Given the prior and transition probabilities of the training data, the total probability of traversing any sequence of $n$ consecutive observations $i_0, \hdots, i_n \in S$ is calculated as
\begin{equation}
Prob(X_0=i_0, \hdots, X_n=i_n)=\rho_{i_0}t_{i_0, i_1}\hdots t_{i_{n-1}, i_{n}}
\end{equation}
For unforeseen states, Laplace smoothing is considered by setting the prior probabilities of and transition probabilities from those states to a tiny value $\delta$.

\subsection{MSHMM Model for PATH}
The proposed Marginally Smoothed Hidden Markov Models (MSHMMs) are specifically trained to handle the unforeseen observations. HMM models are assumed to be generated by a Markov process with unobserved hidden states and are very effective for analyzing sequential data. 
The HMM model can be expressed as $\lambda=(\pi, A, B)$ where the parameters can be learned from the observed locations of the training set $O$ given the vocabulary of possible observations $V$. Here, $\pi$ is the initial hidden-state distribution, $A$ is a time-dependent stochastic transition matrix between the hidden states, and $B$ is a stochastic matrix with the probability of emitting a particular observation at a given state. 

To learn the model, the HMMs are trained using the three-step Baum-Welch algorithm \cite{Rabiner:1990:THM:108235.108253} shown in Algorithm \ref{BaumWelchPath}. The initialization step sets $\lambda=(\pi, A, B)$ with random initial conditions. The parameters are then updated iteratively until convergence.

\begin{algorithm}[h]
\caption{Modified Forward-Backward HMM Algorithm for training MSHMM model for PATH}
\label{BaumWelchPath}
\begin{algorithmic}
\Procedure{BaumWelchForPATH}{training observations $O=\{o_1, o_2, \hdots, o_T\}$ of length $T$, vocabulary of observations $V=\{v_1, v_2, \hdots, v_M\}$ where $M\geq T$}
\State \textbf{initialization} 
\State random initialization of $\pi$, $A$ and $B$. $\delta$.
\While{until convergence}
	\State $\alpha_i(1)=\pi_{i}b_i(o_1)$
	\State $\alpha_i(t+1)=b_i(o_{t+1})\sum_{j=1}^{N}\alpha_j(t)a_{ji} \forall t, i$ 
	\State $\beta_i(T)=1$
	\State $\beta_i(t)=\sum_{j=1}^{N}\beta_j(t+1)a_{ij}b_j(o_{t+1}) \forall t, i$ 
	\State \textbf{E-step}
	\State $\gamma_i(t)=\frac{\alpha_i(t)\beta_i(t)}{\sum_{j=1}^N\alpha_j(t)\beta_j(t)} \forall t, i$ 	
	\State $\xi_{i,j}(t)=\frac{\alpha_i(t)a_{ij}b_j(o_{t+1})\beta_j(t+1)}{\sum_{i=1}^N\sum_{j=1}^N\alpha_i(t)a_{ij}b_j(o_{t+1})\beta_j(t+1)} \forall t, i, j$ 	
	\State \textbf{Modified M-step}
	\State $\widehat{\pi_i}=\gamma_i(1)\forall i$ 
	\State $\widehat{a_{ij}}=\frac{\sum_{t=1}^{T-1}\xi_{i,j}(t)}{\sum_{t=1}^{T-1}\gamma_i(t)}\forall t, i, j$ 
	\If{$\sum_{t=1}^{T}\mathbf{1}(O_t=v_k) \neq 0$}	
		\State $\widehat{b_{i}}(v_k)=\frac{\sum_{t=1}^{T}\mathbf{1}(O_t=v_k)\gamma_i(t)+\delta}{\sum_{t=1}^{T}\gamma_i(t)+T\delta} \forall t, i, j$ 
	\Else
		\State $\widehat{b_{i}}(v_k)=\frac{\sum_{t=1}^{T}\mathbf{1}(O_t^{L,TZ}=v_k^{L,TZ})\gamma_t(j)+\delta}{\sum_{t=1}^{T}\gamma_t(j)+T\delta}\times$\State $\frac{\sum_{t=1}^{T}\mathbf{1}(O_t^{L,W}=v_k^{L,W})\gamma_t(j)+\delta}{\sum_{t=1}^{T}\gamma_t(j)+T\delta} \forall t, i, j$  
	\EndIf
	\State \textbf{Normalize and Update}
	\State update $\pi=normalize(\widehat{\pi})$
	\State update $A=normalize(\widehat{A})$
	\State update $B=normalize(\widehat{B})$
\EndWhile

\State \textbf{return} $\pi$,$A$, $B$
\EndProcedure
\end{algorithmic}
\end{algorithm}

Here $\alpha_i(t)=Prob(y(1)=o_1, \hdots, y(t)=o_t, X(t)=i|\lambda)$ is the probability of seeing the partial observable sequence $o_1,\hdots, o_t$ and ending up in state $i$ at time $t$ and it is calculated recursively. $N$ is the number of hidden states, $a_{ij}$ refers to the $j$-th element of the $i$-row of the $A$ and $b_j(o)$ refers to the emission probabilities of the $j$-th state in $B$ for observation $o$.

The update equation of $\widehat{b_i}$ in the M-step is modified from the original Baum-Welch algorithm to assign non-zero emission probabilities to those observations of the vocabulary $V$ that are not present in the training set $O$. As mentioned in the formulation of the PATH problem, an observation consists of the location, timezone and weekday/weekend information, i.e. $o_i=\{o_i^{L, TZ, WE}\} \forall i \in \{1,\hdots, T\}$. Originally the summation in the nominator of $\widehat{b_i}$ is only made over observed symbols equal to $o_k$, i.e. the indicator function $\mathbf{1}(O_t=v_k)=1$ if $o_t=o_k$, and zero otherwise. But, this equation assigns zero emission probabilities for unforeseen observations, and will eventually pull the overall probability of observing a sequence to zero even if only one such unforeseen yet probable observation is present in the test sequence. One simple work-around is to use a smoothing technique such as Laplace Smoothing where $\widehat{b_i}$ is assigned a very small constant probability. However, the performance of Laplace smoothing is found to be very poor experimentally because of its empirical nature. A marginal smoothing method is proposed here which utilizes the location, timezone and weekend/weekday information of the observations to assign the emission probabilities. The method is based on the assumption that the probability of a user being in a location cluster at a certain timezone $P(v_k^L,v_k^{TZ})$ is independent of the probability of the individual being in a location on weekdays/weekends $P(v_k^L, v_k^W)$. Then, $\widehat{b_i}(v_k)$ can be expressed as

\begin{eqnarray}
\widehat{b_i}(v_k)&=&P(O_t=v_k\mid X_t=i)\nonumber\\
&=&P(O_t^L=v_k^L, O_t^{TZ}=v_k^{TZ}, O_t^W=v_k^W\mid X_t=i)\nonumber\\
&\approx &P(O_t^L=v_k^L, O_t^{TZ}=v_k^{TZ}\mid X_t=i)\nonumber\\
&&\times P(O_t^L=v_k^L, O_t^W=v_k^W\mid X_t=i)\\
&\approx & \frac{\sum_{t=1}^{T}\mathbf{1}(O_t^{L,TZ}=v_k^{L,TZ})\gamma_t(j)}{\sum_{t=1}^{T}\gamma_t(j)}\times \nonumber\\ 
&& \frac{\sum_{t=1}^{T}\mathbf{1}(O_t^{L,W}=v_k^{L,W})\gamma_t(j)}{\sum_{t=1}^{T}\gamma_t(j)}
\end{eqnarray}

where, the indicator function $\mathbf{1}(O_t^{L,TZ}=v_k^{L,TZ})=1$ if the location and timezone of $o_t$ and $v_k$ are the same irrespective of the day and zero otherwise, and $\mathbf{1}(O_t^{L,W}=v_k^{L,W})=1$ only if the location and day are the same irrespective of the timezone and zero otherwise.

If Laplace-smoothing is used instead of the marginal smoothing proposed here, then, when $\sum_{t=1}^{T}\mathbf{1}(O_t=v_k) == 0$, the update equation of $\widehat{b_i}(v_k)$ would be
\begin{equation}
\widehat{b_{i}}(v_k)=\frac{\delta}{\sum_{t=1}^{T}\gamma_i(t)+T\delta} 
\end{equation}
for all t, i. Here, $\delta$ is a very small number, and therefore, a even smaller emission probability is being assigned to an unforeseen observation instead of $0$. 

\section{Experimental Setup and Evaluation}
\begin{figure}[t]
\centering
\includegraphics[width=0.4\textwidth]{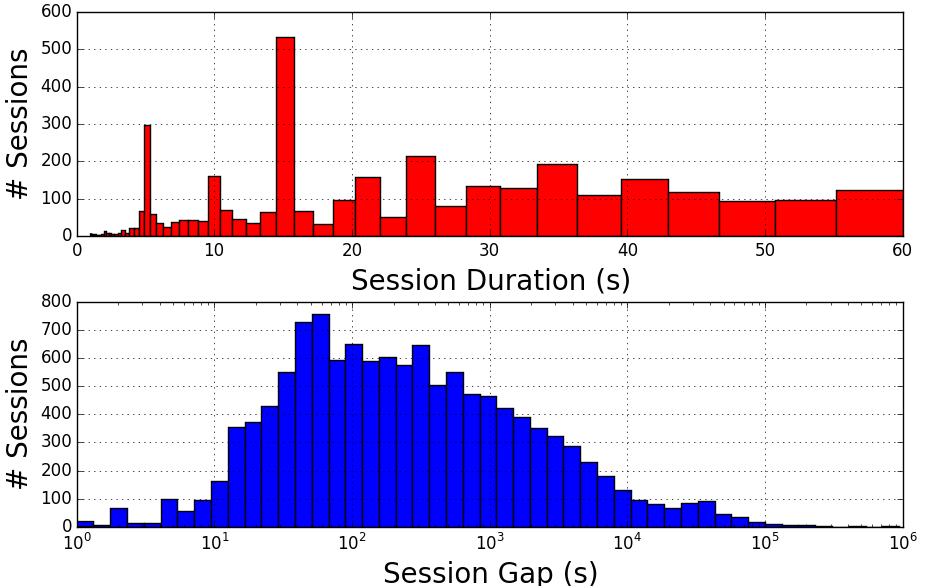}
\vskip 0pt
\caption{For UMDAA02 dataset: (Top) Histogram of duration of sessions and (Bottom) Histogram of time gap between consecutive sessions.}
\label{SessionDurationGap}
\vskip -10pt
\end{figure}

\begin{figure}[t]
\centering
\includegraphics[width=0.4\textwidth]{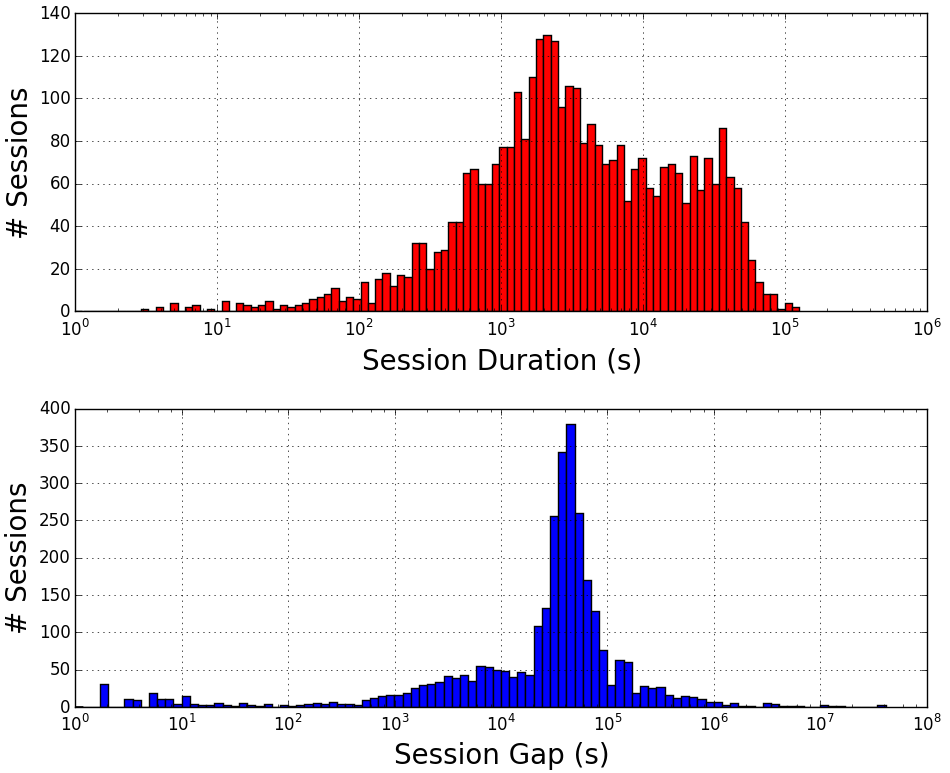}
\vskip 0pt
\caption{For Geo-Life data set: (Top) Histogram of duration of sessions and (Bottom) Histogram of time gap between consecutive sessions.}
\label{SessionDurationGapGeolife}
\vskip -10pt
\end{figure}
Experiments are performed on two datasets: (1) the UMDAA02 geo-location dataset \cite{AA02_MahbubChellappa_BTAS2016}, and (2) the Geolife GPS tarjectory dataset \cite{Geolife_CooperativeSocialNetworkingService}.  The histogram of duration of geo-data collection sessions along with the histogram of time gap between consecutive sessions for the two datasets are shown in Figs. \ref{SessionDurationGap} and \ref{SessionDurationGapGeolife}. It can be seen from these figures that for the UMDAA02 geo-location dataset, geo-location data is collected for at most $60$ seconds after the user logs into the phone. On the other hand, for the Geolife dataset the data is seamlessly collected for long period of times. In fact, the session with the maximum duration is almost $11$ days long. The reason behind this difference is that the UMDAA02 dataset is collected for authentication research using smartphones when the phone is being used, whereas, the GeoLife dataset is collected using GPS-phones and GPS loggers for individual and social behavioral research. Since the event of logging into a phone varies widely, the session gap for the UMDAA02 dataset is spread more widely then the GeoLife dataset.

Considering the nature of the data collection process, experiments for the two methods are designed differently. Since the UMDAA02 dataset is small, sparse and contains user data for a little over a week for each user, the first $70\%$ of chronologically sorted data of each user is used for training the model for that user and the rest are used for evaluation. On the other hand, for the GeoLife dataset, experiments are done on a total of $63$ users who has geo-location data for $6$ weeks or more. Location data for the $6$-th week is used for evaluation, while those from the previous weeks are used for training user-wise models. 

\begin{figure*}[t]
\centering
\subfigure[]{\includegraphics[width = 0.24\textwidth]{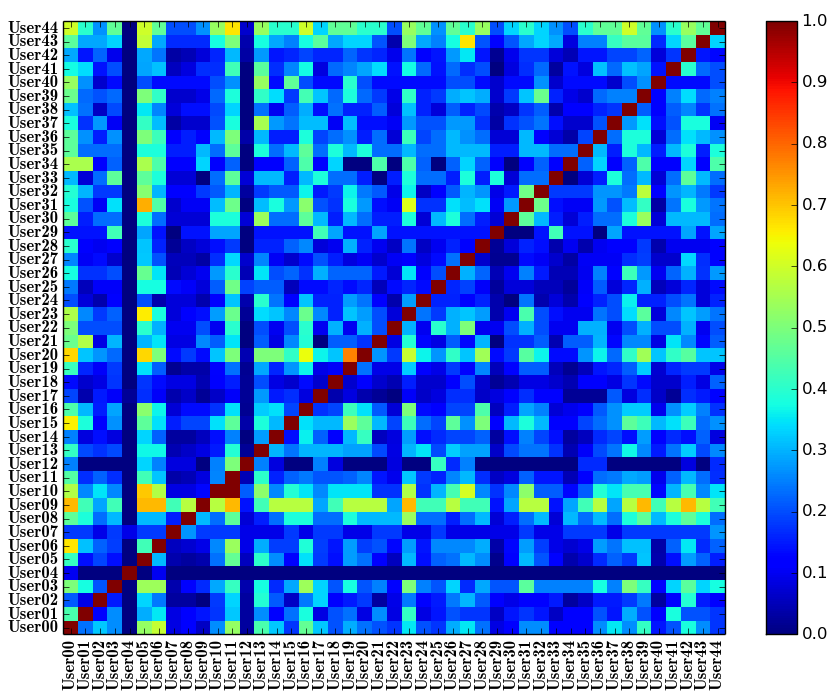}}~
\subfigure[]{\includegraphics[width = 0.245\textwidth]{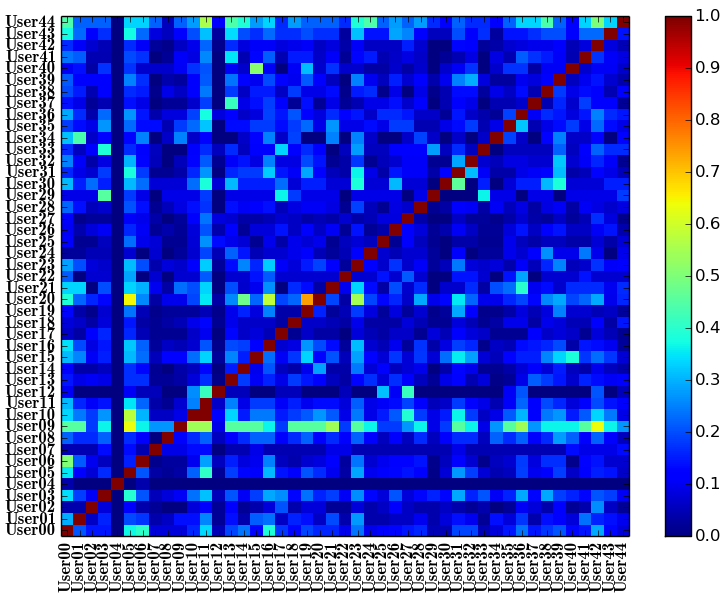}}~
\subfigure[]{\includegraphics[width = 0.233\textwidth]{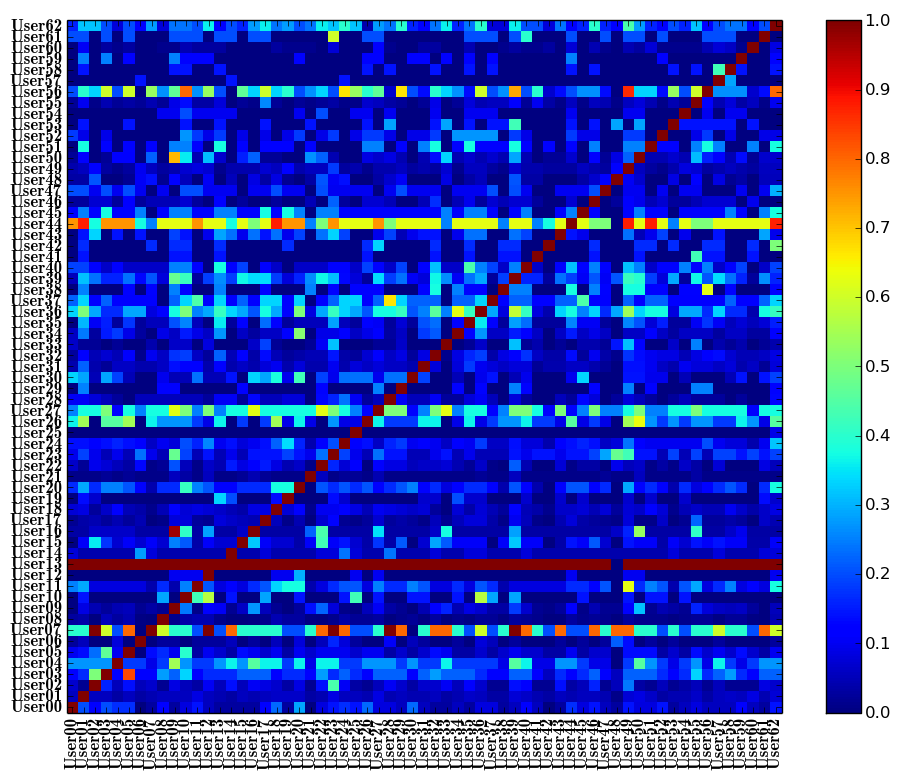}}~
\subfigure[]{\includegraphics[width = 0.25\textwidth]{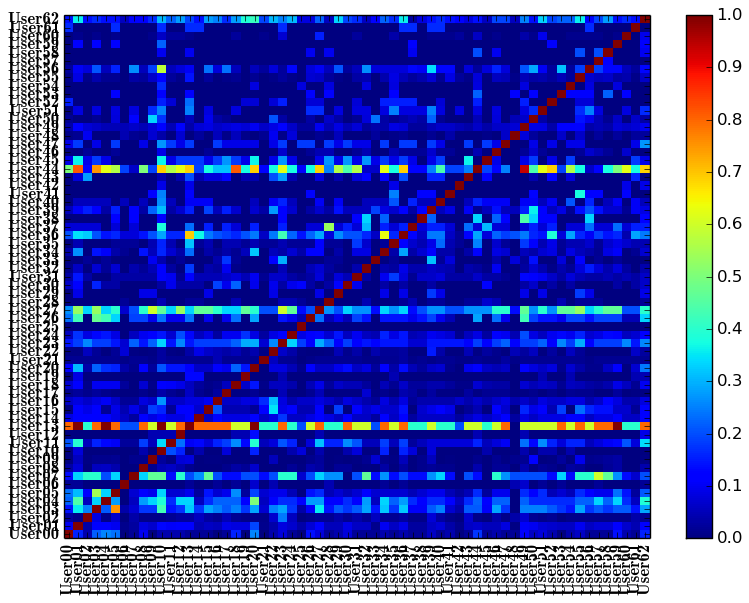}}
\caption{Similarity matrix depicting (a) location overlap for UMDAA02 dataset, (b) observations overlap for UMDAA02 dataset, (c) location overlap for the GeoLife dataset, and, (d) observations overlap for the GeoLife dataset.}
\label{LocationSimilarityMatrix_UMDAA02_GeoLife}
\vskip -10pt
\end{figure*}

In Fig. \ref{LocationSimilarityMatrix_UMDAA02_GeoLife}(a) the distinguishability of location information of the users is depicted as similarity matrices for the UMDAA02 dataset. For a user, after determining the location clusters (considering $R_{max}=20$ meters), the location traces are obtained for the training period of that user and all the other users considering those clusters. In figure \ref{LocationSimilarityMatrix_UMDAA02_GeoLife}(a), the percentage of common location clusters that any two users share is shown as a similarity matrix. It can be seen that in many cases two different user can have significant amount of overlaps. However, when time and day information are incorporated with the location data to generate the observations, the number of overlaps gets reduced, as can be seen from Fig. \ref{LocationSimilarityMatrix_UMDAA02_GeoLife}(b). Intuitively, considering the sequence information would minimize the similarity between two different users even more. Similar depictions for the GeoLife trajectory dataset are shown in Fig. \ref{LocationSimilarityMatrix_UMDAA02_GeoLife}(c) and \ref{LocationSimilarityMatrix_UMDAA02_GeoLife}(d), respectively.

\begin{figure}[t]
\centering
\includegraphics[width=0.4\textwidth]{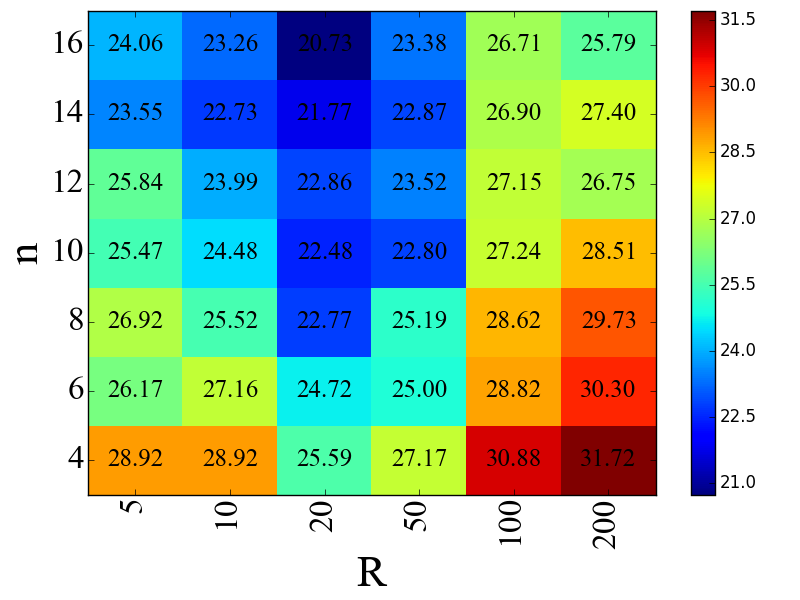}
\vskip 0pt
\caption{EER($\%$) heatmap for length of sequence (n) vs. the cluster maximum radius (R).}
\label{nVsRHeatMap}
\vskip -15pt
\end{figure}

The number of past observations $n$ that is considered when evaluating the verification score has a direct impact on the EER. It can be seen from Fig. \ref{nVsRHeatMap}, that the EER decreases with increasing $n$, which is understandable since having greater number of past observations improves the predictability of the next observation. On the other hand, the EER gets larger if the value of $R$ is too small or too big. The optimum value of $R$ for the UMDAA02 dataset is found to be around $20$ meters, as can be seen in the figure. This is a reasonable value since clusters of this diameter are neither too small to be rooms nor too big to be communities. Rather, they are most likely to be representative of buildings and for the PATH problem and other location-based prediction tasks, generally buildings like home, shopping mall, work-place, gym etc. are considered to translate the geo-location observation.

\begin{figure}[t]
\centering
\includegraphics[width=0.4\textwidth]{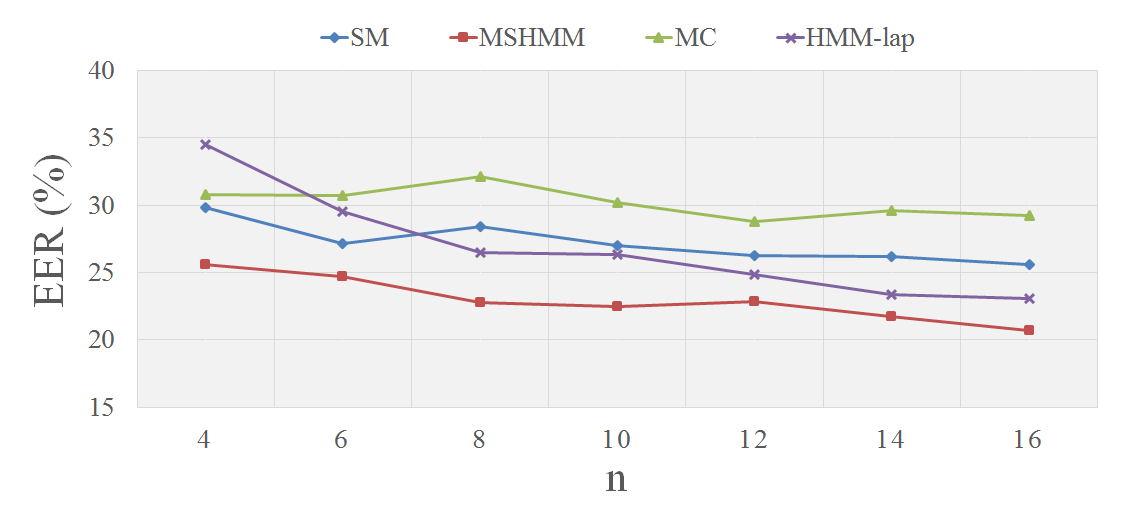}
\vskip 0pt
\caption{Comparison of SM, MC, MSHMM and HMM-lap methods in terms of EER ($\%$) for the UMDAA02 dataset. For HMM-based methods the number of hidden states is $10$.}
\label{EERComparisonUMDAA02}
\vskip -10pt
\end{figure}

In Fig. \ref{EERComparisonUMDAA02}, the performance of the four methods - SM, MSHMM, MC and HMM-lap are shown in terms of EER for varying $n$ on the UMDAA02 dataset for $R_{max}=20$. Given the unconstrained nature of the dataset, verification is a daunting task even with more robust modalities such as face \cite{AA02_MahbubChellappa_BTAS2016}. Yet, the proposed MSHMM method achieved an EER of $20.73\%$ for $n=16$ outperforming all the other methods. In fact, for any value of $n$, the MSHMM models trained with $10$ hidden states performs better then other methods.  

\begin{figure}[t]
\centering
\includegraphics[width=0.4\textwidth]{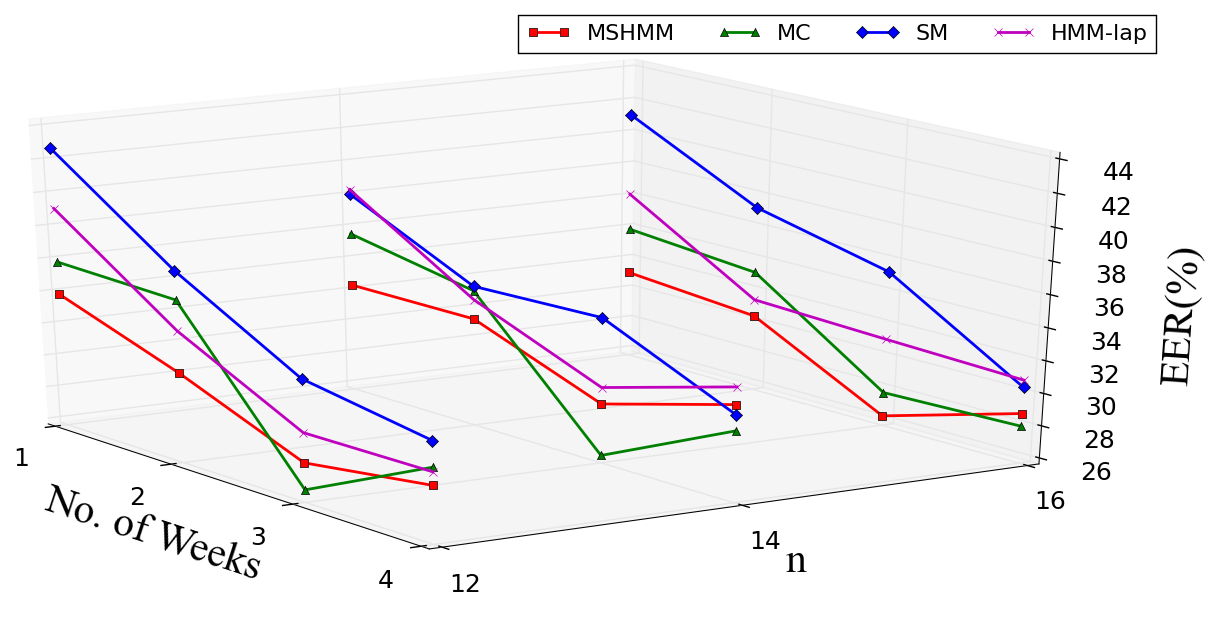}
\vskip 0pt
\caption{Comparison of SM, MC, MSHMM and HMM-lap mehthods in terms of EER ($\%$) for varying $n$ and varying number of weeks of training data. For HMM based methods the number of hidden states is $10$.}
\label{EERComparisonGeoLife}
\vskip -10pt
\end{figure}

Finally, in Fig. \ref{EERComparisonGeoLife}, the performances of the four methods are compared for different $n$.  while the number of past weeks for training is increased from $1$ to $4$. Understandably, the EER is showing a decreasing trend in general for increasing training data. The overall performance of the proposed MSHMM method (trained with $10$ hidden states) is better than the other three methods across different training size and sequence length. The MC method also found to achieve better accuracy on this dataset. This is probably due to the fact that this dataset is less sparse and therefore there are fewer unforeseen observations which led to better estimation of the prior and transition probability matrix for MC. 

\section{Conclusion}
A formulation of location history-based user verification task for active authentication is introduced in this paper. Specifically, information on location clusters, time and day are utilized to obtain observation sequences for users carrying geo-location sensors or smartphones. The probability of a sequence occurring for a particular user is evaluated by training user-specific models. Four different methods, namely, sequence matching (SM), markov chain (MC) and  hidden markov models with marginal smoothing (MSHMM) and Laplace smoothing (HMM-lap), are compared in terms of equal error rate (EER). Through extensive experimentations, it is shown that the proposed MSHMM method outperforms the other methods for two different datasets - UMDAA02 and GeoLife. The future direction of research on this would be to apply similar verification approachs on some other modalities such as wifi and cell data for active authentication and then fusing the scores to obtain even better performance. 

\section*{Acknowledgment}
This work was done in partnership with and supported by Google Advanced Technology and Projects (ATAP), a Skunk Works-inspired team chartered to deliver breakthrough innovations with end-to-end product development based on cutting edge research  and a cooperative agreement FA8750-13-2-0279 from DARPA.

\bibliographystyle{ieee}
\bibliography{LocationBasedVerification_CameraReady}

\end{document}